# CATNet: Cross-event Attention-based Time-aware Network for Medical Event Prediction

Sicen Liu[1], Xiaolong Wang[1], Yang Xiang[2], Hui Xu[3], Hui Wang[3], Buzhou Tang[1,2,*]

[1]Department of Computer Science, Harbin Institute of Technology Shenzhen Graduate School, Shenzhen, China
[2]Pengcheng Laboratory, Shenzhen, China.
[3]Gennlife (Beijing) Technology Co Ltd, Beijing, China.

Corresponding Author: Buzhou Tang, Ph.D., Department of Computer Science, Harbin Institute of Technology Shenzhen Graduate School, Shenzhen, China, 518055. Email: tangbuzhou@gmail.com

**Abstract**

Medical event prediction (MEP) is a fundamental task in the medical domain, which needs to predict medical events, including medications, diagnosis codes, laboratory tests, procedures, outcomes, and so on, according to historical medical records. The task is challenging as medical data is a type of complex time series data with heterogeneous and temporal irregular characteristics. Many machine learning methods that consider the two characteristics have been proposed for medical event prediction. However, most of them consider the two characteristics separately and ignore the correlations among different types of medical events, especially relations between historical medical events and target medical events. In this paper, we propose a novel neural network based on attention mechanism, called cross-event attention-based time-aware network (CATNet), for medical event prediction. It is a time-aware, event-aware and task-adaptive method with the following advantages: 1) modeling heterogeneous information and temporal information in a unified way and considering temporal irregular characteristics locally and globally respectively, 2) taking full advantage of correlations among different types of events via cross-event attention. Experiments on two public datasets (MIMIC-III and eICU) show CATNet can be adaptive with different MEP tasks and outperforms other state-of-the-art methods on various MEP tasks. The source code of CATNet will be released after this manuscript is accepted.

## 1. Introduction

Nowadays, with the development of electronic medical record (EMR) systems, massive clinical data about patients is available for medical research and healthcare [1–8]. Many machine learning methods, especially deep learning methods, have been used for medical event prediction (MEP) [9–15] (as shown in Fig. 1) to support clinical decisions. Formally, given historical medical records of a patient p with demographic information $S$ and $T$ visits, each of which $v_t$ includes various kinds of medical events (usually denoted by medical codes) such as diagnoses $x_t^d$, procedures $x_t^p$, laboratory tests (labtests) $x_t^l$, medications $x_t^m$ and other medical events $x_t^?$, the goal of medical event prediction is to predict the medical events at time T+1, denoted by $y_{T+1}$. The medical events at time T+1 may be diagnoses, procedures, labtests, medications or other medical events that are of the same types as the medical events in the historical records, or of new types different from the medical events in the historical records. According to the type of target events, MEP can be classified into risk prediction [16–18], patient treatment trajectory [19,20], readmission prediction [21–24], mortality prediction [17,25–28], diagnosis code prediction [29–32], prescription prediction [33,34] etc.

---
[*] corresponding author

These MEP tasks usually face the following three challenges:

1) Information decay rates with time are related to temporal irregular intervals between two neighbor visits (i.e., $\Delta t_t$) as well as medical events. The same medical events with irregular temporal intervals should correspond to different decay rates, and the different medical events should be decayed in different ways.

2) There are some correlations among medical events in a visit, which should be considered appropriately. For instance, when a patient with acute renal failure and his/her potassium values exceed the safe limit in a laboratory test, the physician will prescribe Furosemide, Calcium Gluconate, and Potassium Chloride to the patient.

3) If the medical events in historical records are of the same type as target medical events, these types of medical events should be paid more attention than other types of medical events. Methods for MEP should be task-adaptive.

Most machine learning methods focus on solving one or two of the three challenges. For example, T-LSTM [35] is a time-aware LSTM (long short-term memory network) designed to incorporate the temporal interval between two neighbor visits into the memory cell of the basic LSTM unit to adjust the information transmission from the previous visit to the next visit. It improves the performance of the standard LSTM. HiTANet [18] is a hierarchical time-aware attention network based on Transformer, which models temporal information in local and global stages to imitate the decision-making process of doctors in risk prediction.

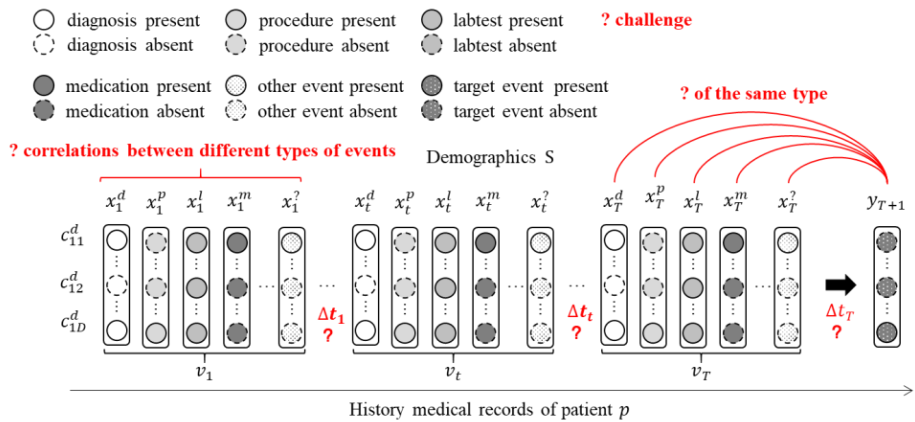

**Fig. 1** Medical event prediction and its challenges.

HiTANet achieves much better performance than T-LSTM on three disease prediction tasks. Using the same structure of T-LSTM, the heterogeneous LSTM (LSTM-DE) [33] adds a gate into the basic LSTM unit to model potential relations between two types of heterogeneous information, where the medical events of the same types as the target events are primary events and the other medical events of different types from the target events are auxiliary events. RetainEx [36] is an attention-based recurrent neural network (RNN) for risk prediction, which learns event-level weights within a visit as well as visit-level weights for interpretability. The attention mechanism is used to model correlations among medical events in a visit.

In addition, RetainEx also considers the irregular temporal intervals by incorporating visit temporal intervals into the input of RNN.

To tackle all the three challenges mentioned above, we design a novel neural network based on a novel attention mechanism, called cross-event attention-based time-aware network (CATNet), for medical event prediction. In CATNet, we regard temporal intervals as medical events a "new" type and design a novel attention mechanism (called cross-event attention) to model correlations among medical events in the same visit, including temporal intervals. The cross-event attention has two forms corresponding to the type of target medical events to make CATNet task-adaptive. Moreover, in order to model temporal information in a global aspect, CATNet also introduces a global time convertor to consider the time medical events decay as impact factors for MEP.

In summary, the proposed CATNet method has the following contributions:

- We design a time-aware, event-aware, and task-adaptive network to model heterogeneous and irregular time series medical data.

- We regard temporal information as a "new" type of medical event and propose a novel attention mechanism (cross-event attention) to learn information decay rate for each visit and correlations among medical events in each visit in a unified way. This attention mechanism is time-aware and task-adaptive.

- We introduce a visit-level attention to model the relations among historical visits, and a global time convertor to model temporal information globally.

- CATNet can be used as a universal framework for MEP. We conduct experiments on two public real-world datasets to verify the effectiveness of CATNet we propose. Ablation studies and result analysis show the effectiveness of the proposed method.

## 2. Related work

In recent years, various deep learning methods have been proposed for medical event prediction on EMR data, including multi-layer perception (MLP) [37], convolutional neural network (CNN) [20,38,39], RNN [21,26,34,40] and Transformer [18] [41] . Among them, RNN is the representative one, and Transformer also shows great potential. We review these studies from three perspectives corresponding to the three challenges mentioned above.

**Time-aware Deep Learning Methods for Medical Event Prediction**. Most of the time-aware deep learning methods are extended from basic RNN, GRU (gated recurrent unit network), or LSTM that is suitable for sequential data with identical temporal intervals. T-LSTM [35] uses a decay function of the temporal interval to control information transmission between neighbor visits and integrates the function into the memory cell of the basic LSTM unit. It achieves much better performance than LSTM on risk prediction. StageNet [17] is a stage-aware LSTM that integrates temporal intervals into the basic LSTM cell to represent the disease progression stage. Timeline [41] applies a time-aware disease progression function to determine how much disease information is transmitted from the current visit to the target visit before inputting visit representation into LSTM. DATA-GRU [11] and RetainEx [36] incorporate visit temporal intervals as additional features into the input vectors of RNN. Besides RNN and LSTM extensions, the attention mechanism [42], [43], and time-aware graph neural networks [34] are also used to model irregular temporal intervals. ConCare [44] applies time-aware attention to the output of GRU. HiTANet [18], a Transformer-based method, uses hierarchical time-aware attention to model temporal information at local and global levels. RGNN_TG_ATT [34] connects medical events in neighbor visits with edges weighted by temporal intervals to form a time-aware graph.

**Event-aware Deep Learning Methods for Medical Event Prediction**. Some studies try to investigate medical event prediction at both visit-level and event-level. RetainEx [36] adopts an attention mechanism to model the correlations among medical events in a visit. ConCare [44] uses GRU to model each event sequence.

**Task-adaptive Deep Learning Methods for Medical Event Prediction**. Task-adaptive methods consider the types of target events. They assume that medical events in historical records of the same type as target events should be paid more attention to. LSTM-DE [33] proposed for medication prediction considers medications as primary events and labtests as auxiliary events and adopts a similar network with T-LSTM to emphasize the primary events by introducing a decomposition structure into LSTM. RGNN_TG_ATT extends LSTM-DE by introducing an additional time-aware graph neural network. DA-LSTM [45] is an extension of T-LSTM that uses a context fusion module to enhance auxiliary inputs by considering current inputs and whole input sequences instead of the decomposition structure in T-LSTM.

Most existing deep learning methods focus on one or two challenges of MEP. The proposed CATNet considers the three challenges of MEP in a unified framework comprehensively.

## 3. Materials and method

### 3.1 *Task definition*

This section defines some notations and describes the MEP tasks.

**Definition 1 (Medical Events)**. Medical events are events related to healthcare in EMRs, including medications, diagnoses, labtests, procedures, mortality status, etc. They are usually normalized by different types of codes. In this study, we only consider five types of medical events, that is, medications, diagnoses, labtests, procedures, and mortality status, each of them is denoted by $E^{event}, event \in \{m, d, l, p, mo\}$, where m, d, l, p, and mo denote medication, diagnosis, labtest, procedure, and mortality respectively. Suppose that there are $M$ codes in $E^m$, $D$ codes in $E^d$, $L$ codes in $E^l$ and $P$ codes in $E^p$.

**Definition 2 (Binary EMR Data)**. A patient $p$ with $T$ historical visits $(v_1, v_2, \cdots, v_T)$ and demographics $S$ can be represented by a three-dimensional binary matrix as shown in Figure 1, where $x_i^{event}$ denotes the event type of events in the $i$-th visit $v_i$, $c_{ik}^{event}$ indicates whether the $k$-th event code is present in the $i$-th visit if the $k$-th event code appears in the $i$-th visit, $c_{ik}^{event} = 1$, otherwise $c_{ik}^{event} = 0$.

**Definition 3 (Temporal Interval)**. Let $t_i$ denote the corresponding time of the $i$-th visit $v_i$. Let $x_i^{\Delta t} = t_{i+1} - t_i$ represent the time interval between the ($i$+1)-th visit and the i-th visit. Let $t_{g_i}$ denote the corresponding time of the $i$-th visit $v_i$ which compares with the first visit record of the patient. Let $x_i^{\Delta gt} = t_{g_i} - t_{g_1}$ represent the temporal gap between the current visit record and the first visit record of the patient.

**Problem (Medical Event Prediction).** Given a patient with $T$ visits and demographics $S$, $p = [v_1, v_2, \cdots, v_T, v_*, S]$, the goal of medical event prediction is to predict which medical events will appear in the next visit. In this study, we investigate the following two cases: 1) the historical visits include the medical events of the same type as the target events. For example, medication prediction according to medications, diagnoses, labtests, and procedures in the historical visits. 2) the target events are of new types different from all medical events in the historical visits. For example, mortality prediction according to medications, diagnoses, labtests, and procedures in the historical visits.

### 3.2 *Method*

This section presents our proposed method (as shown in Fig. 2). CATNet consists of seven components: raw visit representation, visit embedding representation, cross-event attention, visit sequence modeling network, local attention, global time convertor, and event prediction. Given a raw visit vector $[x_t^m, x_t^d, x_t^l, x_t^p] = [c_{t1}^m, \ldots, c_{tM}^m, c_{t1}^d, \ldots, c_{tD}^d, c_{t1}^l, \ldots, c_{tL}^l, c_{t1}^P, \ldots, c_{tP}^p]$, each code $c_{tk}^{event}$ is embedded into a dense vector $e_{tk}^{event} \in R^n$. To take full advantage of temporal information, using the similar method with [18], we embed each temporal interval $x_t^{\Delta t}$ (or $x_t^{\Delta gt}$) into a vector as follows:

$$e_t^{\Delta t} = W_{\Delta t2}\left(1 - \tanh\left(\left(W_{\Delta t1} x_t^{\Delta t} + b_{\Delta t1}\right)^2\right)\right) + b_{t2}, \qquad (1)$$

where $W_{\Delta t1} \in R^a$, $b_{\Delta t1} \in R^a$, $W_{\Delta t2} \in R^{n \times a}$ and $b_{\Delta t2} \in R^n$ are all parameters. Thus, the visit $v_t$ with temporal interval $x_t^{\Delta t}$ can be represented by $e_t = [e_t^m, e_t^d, e_t^l, e_t^p, e_t^{\Delta t}] = [e_{t1}^m, \ldots, e_{tM}^m, e_{t1}^d, \ldots, e_{tD}^d, e_{t1}^l, \ldots, e_{tL}^l, e_{t1}^p, \ldots, e_{tP}^p, e_t^{\Delta t}]$, where the temporal information is regarded as the same as medical events. In this way, temporal intervals and medical events are represented in a unified way.

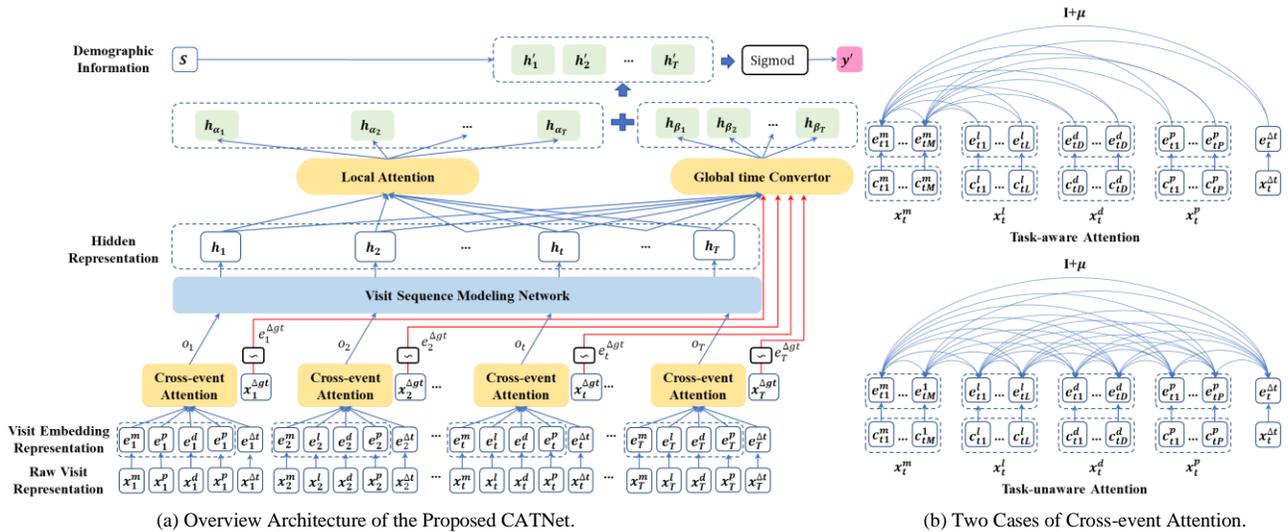

(a) Overview Architecture of the Proposed CATNet.      (b) Two Cases of Cross-event Attention.

**Fig. 2** The architecture of CATNet is shown in Figure2(a). Figure 2(b) gives two cases of cross-event attention, the upper is an example of task-unware attention, where the task is mortality prediction according to medications, diagnoses, labtests, and procedures in historical visits, the lower is an example of task-aware attention, where the task is medication prediction with medications as primary events and others as auxiliary events.

For visit embeddings $e_t$, we design the cross-event attention and apply it on them to obtain time-aware, event-aware, and task-adaptive visit embeddings $o_t$ (For details, refer to the cross-event attention section). We further input the new visit embeddings into a common sequence modeling network (e.g, GRU, LSTM, or Transformer) to obtain the hidden representation of each visit $h_t$. Visit-level attention and global time convertor are applied to $h_t$ and $x_t^{\Delta gt}$ to utilize temporal information as well as historical medical events at visit levels and global levels respectively. The representation obtained by visit-level attention $h_{\alpha_t}$ and the representation obtained by the global time convertor $h_{\beta_t}$ are combined together. We also consider the demographic information of the patient as the static characteristic. Finally, a fully connected network with the sigmoid function is used for prediction. In the following sections, we present cross-event attention, local attention, global time convertor, and event prediction in detail.

*3.2.1 Cross-Event Attention*

There are two cases of cross-event attention according to the type of the target events: task-unware attention and task-aware attention. The task-unware attention corresponds to the case that the target events are of new types different from all events in the historical visits, while the task-aware attention corresponds to the case that the historical visits include the medical events of the same type as the target events (i.e, primary events). We apply self-attention on only primary events (take medication prediction as an example) in the task-aware attention (Eq. (2)), but all events in the task-unware attention (Eq. (3)) as follows:

$$\mu_{ti}^m = softmax(\frac{e_{ti}^m (\hat{e}_{ti}^{event})^T}{\sqrt{n}}), \tag{2}$$

$$\mu_{ti}^{event} = softmax(\frac{e_{ti}^{event} (\hat{e}_{ti}^{event})^T}{\sqrt{n}}), \tag{3}$$

where $\hat{e}_{ti}^{event} \in R^{(M+D+L+P+1) \times n}$ is a matrix of all event and temporal embeddings.

The output of cross-event attention is set as:

$$o_t = \begin{cases} [o_{t1}^m, \dots, o_{tM}^m, o_t^{\Delta t}] \ (task-aware) \\ [o_{t1}^m \dots, o_{t1}^d \dots, o_{t1}^l \dots, o_{t1}^p \dots, o_{tP}^p, o_t^{\Delta t}](task-unware) \end{cases}, \tag{4}$$

$$o_{ti}^{event} = \begin{cases} (e_{t1}^m + \mu_{t1}^m \hat{e}_{t1}^{event})(task-aware) \\ (e_{ti}^{event} + \mu_{ti}^{event} \hat{e}_{ti}^{event}) \ (task-unware) \end{cases} \tag{5}$$

Because of the attention weights between the primary events (or all events) and temporal intervals, our proposed CATNet is time-aware. As the attention is applied at event level, CATNet is event-aware. Moreover, the attention mechanism can be adaptive to different tasks.

Therefore, we obtain time-aware, event-aware, and task-adaptive visit embeddings via cross-event attention. For convenience, the CATNet using task-unware attention is denoted by CATNet-I, and that using task-aware attention is denoted by CATNet-II. CATNet-II can be regarded as a specific version of CATNet-I.

*3.2.2 Visit-level Attention*

Any sequence modeling network can be used as a backbone network to model historical visit sequences, such as GRU, LSTM, and Transformer. Suppose that the hidden representation of the time-aware, event-aware, and task-adaptive visit embedding sequence $[o_1, o_2, \dots, o_T]$ can be obtained via the following equation:

$$h = [h_1, h_2, \dots, h_T] = Backbone[o_1, o_2, \dots, o_T], \tag{6}$$

where $h_t \in R^b$ is the hidden state for the t-th visit by aggregating all the medical information, and Backbone is a sequence modeling network.

After obtaining $h$, we employ visit-level attention to generate attention weights for each visit as follows:

$$[\alpha_1, \ldots, \alpha_T] = [\text{softmax}\left(\frac{h_1 h_{1:T}^T}{\sqrt{b}}\right), \ldots, \text{softmax}\left(\frac{h_T h_{1:T}^T}{\sqrt{b}}\right)], \quad (7)$$

where $h_{1:T}$ is a matrix of hidden states for visits from 1 to T. Then a patient can be represented as follows:

$$h_{\alpha_t} = \sum_{t=1}^{T} \alpha_t h_t \quad (8)$$

The local attention module can capture the local relations among historical visits.

### 3.2.3 Global time Convertor

Besides temporal intervals between neighbor visits, we also consider the times medical events decay. A global time convertor is designed to generate the impact score of each medical decay time globally. Similar to Eq.(1), a global medical event decay time $x_t^{\Delta gt}$ can be embedded as:

$$e_t^{\Delta gt} = W_{\Delta g_{t2}}\left(1 - \tanh\left(\left(W_{\Delta g_{t1}} x_t^{\Delta gt} + b_{\Delta g_{t1}}\right)^2\right)\right) + b_{\Delta g_{t2}}, \quad (9)$$

where $W_{\Delta g_{t1}} \in R^b$, $b_{\Delta g_{t1}} \in R^b$, $W_{\Delta g_{t2}} \in R^{m \times b}$ and $b_{\Delta g_{t2}} \in R^m$ are all parameters. Furthermore, we adopt the sigmoid activation function to obtain the impact score of each medical decay time global as follows:

$$[\beta_1, \ldots, \beta_T] = \sigma([e_1^{\Delta gt}, \cdots, e_T^{\Delta gt}]), \quad (10)$$

where σ is the sigmoid activation function. Then we obtain a patient representation considering temporal information globally:

$$h_{\beta_t} = \sum_{t=1}^{T} \beta_t h_t \quad (11)$$

### 3.2.4 Event Prediction

For event prediction, we consider the patient representation obtained by the local attention module, the patient representation obtained by the global time convertor as well as the patient's demographic information S. The two patient representations are first added together via $h^{'} = sum([h_1, h_2, \cdots h_T])$, where $h_t = h_{\alpha_t} + h_{\beta_t}$. Then $h^{'}$ and is concatenated with the representation of the patient's demographic information S to form the final input of the event prediction module, that is, $[h^{'}, e^s]$.

To obtain the representation of the patient's demographic information S, we apply a fully connected network to S:

$$e^s = (W_s S + b_s), \quad (12)$$

where $W_s \in R^{s \times g}$ and $b_s \in R^s$ are parameters.

Finally, a fully connected network with the sigmoid function is used to make a binary (vector) prediction as follows:

$$y' = \sigma(W_u[h', e^s] + b_u), \quad (13)$$

where $W_u \in R^{\rho \times (b+g)}$ and $b_u \in \mathbb{R}^\rho$ are parameters, and $\rho$ is the total number of events that need to be predicted. Binary cross-entropy is used as the loss function for each event.

**4. Experiments results**

We conduct experiments on two public real-world datasets, i.e., MIMIC-III [46] and eICU Collaborative Research Dataset [47] to evaluate the performance of the proposed CATNet.

*4.1 Description of Datasets*

**MIMIC-III**[1]: MIMIC-III (Medical Information Mart for Intensive Care III) is a freely available database comprising de-identified medical records of over 58K patients who stayed in the ICUs at the Beth Israel Deaconess Medical Center between 2001 and 2012. It contains various types of medical events, such as medications, diagnoses, labtests, procedures, and mortality status. Following the previous work [34], we select the patients with at least two visits and drop medications with occurrence frequency lower than 2000. Finally, we obtain 5438 patients for medication prediction, diagnosis prediction, labtest prediction, procedure prediction, and mortality prediction. We apply CATNet with task-aware attention to the former four-event prediction tasks and CATNet with task-unware attention to mortality prediction.

**eICU**[2]: The eICU collaborative research database is collected from 208 critical care units in the continental United States. The data contains the medical records of over 200K patients admitted to ICUs between 2014 and 2015. We also select the patients with at least two visits for experiments. Three types of medical events, that is, medications, labtest, and procedures, are used for event prediction. Finally, we obtain 9215 patients and apply CATNet with task-aware attention to medication prediction, labtest prediction respectively.

The statistics details of the selected data in the two datasets are summarized in Table 1, where "# *" denotes the number of *, "Avg" and "Max" are abbreviations of "averaged" and "maximum". In this study, we split each dataset into training, validation, and test sets across users with ratios of 8:1:1 in all experiments.

**Table 1**
Statistics of the datasets

| Dataset | MIMIC-III | eICU |
|---|---|---|
| #Patients | 5438 | 9215 |
| Unique #Diagnosis codes | 905 | 1366 |
| Unique #Procedure codes | 529 | NA |
| Unique #Medication codes | 346 | 1375 |
| Unique #Labtest codes | 676 | 155 |
| #mortality status = '1' | 1063 | NA |
| Avg(Max) #visits per patient | 2.59(29) | 2.19(9) |
| Avg(Max) #diagnosis codes per visit | 12.92(87) | 8.22(106) |
| Avg(Max) #procedure codes per visit | 4.12(37) | NA |
| Avg(Max) #medication codes per visit | 32.00(172) | 14.73(100) |
| Avg(Max) #labtest codes per visit | 71.26(252) | 38.36(87) |

*4.2 Experimental Settings*

We compare CATNet with the following state-of-the-art methods:

---
[1] https://mimic.physionet.org
[2] https://eicu-crd.mit.edu/about/eicu/

1) **DoctorAI** [9]: This method is a classical medical event prediction method that uses the basic RNN to represent patient historical visits.

2) **T-LSTM** [35]: This method is a time-aware LSTM designed to incorporate the temporal interval between two neighbor visits into the memory cell of the basic LSTM unit to control the information transmission.

3) **LSTM-DE** [33]: This method is a task-aware LSTM that introduces a gate into the basic LSTM unit to model potential correlations between two types of medical events. It has a similar structure as T-LSTM.

4) **RGNN_TG_ATT** [34]: This method is a hybrid method of LSTM-DE and GNN (graph neural network) to represent patient visit sequences, where a time-aware GNN is used to capture global information of temporal medical events.

5) **RetainVis** [36]: This method is an RNN-based method that integrates visit temporal intervals as additional features to the input vectors of RNN and applies attention to capture the correlations among medical events in a visit.

6) **StageNet** [17]: This method is a stage-aware LSTM that pays attention to the disease progression stage by integrating temporal intervals into the basic LSTM unit and using CNN to model correlations between neighbor stages.

7) **ConCare** [44]: This method uses GRU to model each event sequence and applies time-aware attention to the output of GRU.

8) **HiTANet** [18]: This method is a hierarchical time-aware Transformer-based method that uses hierarchical time-aware attention to utilize temporal information at local and global levels.

They fall into two categories: RNN variants (1-7) and Transformer variants (8). It should be noted that among the eight baseline methods, LSTM-DE and RGNN_TG_ATT cannot be directly applied to mortality prediction as they are task-adaptive, and we only apply them on medication prediction as they can only consider one auxiliary type of medical events. In addition, we can only conduct mortality experiments on the MIMIC-III dataset as only it contains mortality status. We implement CATNet in the PyTorch framework and use the source codes of DoctorAI[3], T-LSTM[4], ReTainEx[5], LSTM-ED[6], RGNN_TG_ATT[7], StageNet[8], ConCare[9], and HiTANet[10] as their implementations. For all the methods, we train the models with randomly initialized parameters 10 times on the training sets For all the methods, we train the models with randomly initialized parameters 10 times on the training sets, save the best models on the validating datasets at each time, and test the best models on the test sets. The mean performances of the 10 times independent runs on the test datasets and their standard deviations are reported in the "Experimental and Analysis" section.

4.3 *Metrics*

Three metrics: the area under the receiver operating characteristic curve (AUC), the area under the precision-recall curve (AUPR), and Top-*k* recall are used as performance evaluation metrics. During model training, we set epochs as 100 epochs, the learning rate as 0.0001, the hidden size of CATNet on the MIMIC-III dataset as 256, and 512 on the eICU dataset, the dropout rate as 0.3, and the other hyperparameters as default.

---

[3] https://github.com/mp2893/doctorai
[4] https://github.com/illidanlab/T-LSTM
[5] https://github.com/mp2893/retain,
[6] The implementation is provided by the authors of [33]
[7] The implementation is provided by the authors of [18]
[8] https://github.com/v1xerunt/StageNet
[9] https://github.com/Accountable-Machine-Intelligence/ConCare
[10] https://github.com/HiTANet2020/HiTANet

4.4 *Experimental Results and Analysis*

The comparisons of CATNet with the other methods for medicine prediction, diagnosis prediction, labtest prediction, and procedure prediction on the two datasets in AUC and AUPR are reported in Table 2 and Table 3, where "NA" denotes no result. The comparisons of CATNet with other methods for mortality prediction on the MIMIC-III dataset in AUC and AUPR are reported in Table 4. By analyzing the results, the following conclusions could be summarized.

**Table 2**
Auc and Aupr of different methods on medication prediction, diagnosis prediction, procedure prediction, and labtest prediction on the MIMIC-III dataset

| MIMIC_III | model | medication | | diagnosis | | procedure | | labtest | |
|---|---|---|---|---|---|---|---|---|---|
| | | Auc(%) | Aupr(%) | Auc(%) | Aupr(%) | Auc(%) | Aupr(%) | Auc(%) | Aupr(%) |
| RNN | DoctorAI | 84.15±0.16 | 35.69±0.26 | 93.83±0.02 | 25.37±0.29 | 94.16±0.07 | 24.23±0.27 | 96.92±0.03 | 82.36±0.38 |
| | T_LSTM | 79.72±0.65 | 29.77±0.97 | 91.71±0.41 | 24.89±1.04 | 90.85±0.76 | 19.13±1.04 | 96.00±0.22 | 81.33±0.73 |
| | LSTM-ED | 84.08±0.17 | 36.70±0.15 | NA | NA | NA | NA | NA | NA |
| | RGNN_TG_ATT | 85.06±0.34 | 39.78±0.61 | NA | NA | NA | NA | NA | NA |
| | RetainVis | 82.09±0.13 | 33.69±0.39 | 92.85±0.06 | 28.22±0.24 | 92.82±0.26 | 22.48±0.64 | 96.71±0.03 | 83.47±0.11 |
| | StageNet | 83.93±0.20 | 36.26±0.37 | 94.03±0.04 | 30.27±0.41 | 94.05±0.12 | 23.12±1.14 | 97.20±0.03 | 84.21±0.15 |
| | ConCare | 75.37±0.39 | 27.32±0.26 | 94.57±0.11 | 29.34±0.70 | 93.73±0.04 | 23.65±0.22 | 97.07±0.01 | 84.41±0.02 |
| | **CATNet_I(GRU)** | 85.40±0.12 | 39.21±0.27 | 94.65±0.05 | 38.59±0.32 | 94.43±0.09 | 28.41±0.69 | 97.29±0.02 | 85.37±0.12 |
| | **CATNet_I(LSTM)** | 85.41±0.12 | 39.36±0.22 | 94.68±0.02 | 38.77±0.27 | 94.42±0.15 | 27.78±0.72 | 97.30±0.03 | 85.42±0.11 |
| | **CATNet_II(GRU)** | **85.77±0.10** | **40.03±0.31** | **95.00±0.03** | **40.96±0.15** | **94.59±0.10** | **29.39±0.44** | **97.31±0.02** | **85.46±0.10** |
| | **CATNet_II(LSTM)** | **85.84±0.10** | **40.23±0.28** | **94.95±0.06** | **40.41±0.24** | **94.59±0.09** | **29.27±0.44** | **97.33±0.02** | **85.60±0.10** |
| Transformer | HiTANet | 83.48±0.13 | 37.09±0.26 | 94.59±0.04 | 38.91±0.20 | 94.48±0.07 | 28.48±0.41 | 97.18±0.02 | 85.00±0.15 |
| | **CATNet_I(Transformer)** | 85.82±0.09 | 39.94±0.24 | 94.85±0.05 | 39.60±0.26 | 94.61±0.09 | 29.37±0.41 | 97.34±0.01 | 85.52±0.06 |
| | **CATNet_II(Transformer)** | **86.21±0.04** | **40.78±0.17** | **95.17±0.03** | **42.45±0.23** | **94.67±0.07** | **30.00±0.24** | **97.40±0.01** | **85.71±0.07** |

**Table 3**
Auc and Aupr of different methods on medication prediction, diagnosis prediction, procedure prediction, and labtest prediction on the eICU dataset

| eICU | model | medication | | diagnosis | | procedure | | labtest | |
|---|---|---|---|---|---|---|---|---|---|
| | | Auc(%) | Aupr(%) | Auc(%) | Aupr(%) | Auc(%) | Aupr(%) | Auc(%) | Aupr(%) |
| RNN | DoctorAI | 92.94±1.14 | 13.13±2.02 | 93.68±0.34 | 22.96±2.96 | NA | NA | 95.16±0.02 | 84.31±0.22 |
| | T_LSTM | 89.29±1.14 | 17.56±1.06 | 92.78±0.07 | 22.24±1.41 | NA | NA | 94.80±0.11 | 85.19±0.28 |
| | LSTM-ED | 95.47±5.52 | 22.58±6.17 | NA | NA | NA | NA | NA | NA |
| | RGNN_TG_ATT | 97.74±0.04 | 27.43±0.38 | NA | NA | NA | NA | NA | NA |
| | RetainVis | 96.65±0.06 | 24.27±0.31 | 93.48±0.08 | 37.01±1.18 | NA | NA | 95.54±0.04 | 86.95±0.11 |
| | StageNet | 97.78±0.07 | 31.13±0.55 | 94.94±0.14 | 42.17±1.57 | NA | NA | 95.79±0.05 | 87.80±0.14 |
| | ConCare | 96.41±0.21 | 30.31±1.13 | 94.36±0.35 | 25.27±3.21 | NA | NA | 94.73±0.25 | 88.19±0.56 |
| | **CATNet_I(GRU)** | 97.65±0.03 | 30.62±0.54 | **96.06±0.08** | 57.84±0.34 | NA | NA | 96.02±0.05 | 88.02±0.22 |
| | **CATNet_I(LSTM)** | 97.65±0.04 | 30.65±0.60 | 96.01±0.09 | **58.00±0.26** | **NA** | **NA** | 96.10±0.03 | 88.24±0.17 |
| | **CATNet_II(GRU)** | 97.90±0.03 | 33.18±0.34 | 96.24±0.08 | 61.00±0.16 | NA | NA | 96.13±0.05 | 88.32±0.20 |
| | **CATNet_II(LSTM)** | **97.96±0.04** | 32.81±0.47 | **96.14±0.10** | 60.41±0.29 | **NA** | **NA** | 96.13±0.03 | 88.28±0.13 |
| Transformer | HiTANet | 97.32±0.38 | 29.08±0.80 | 96.15±0.41 | 54.41±2.75 | NA | NA | 96.01±0.03 | 88.18±0.13 |
| | **CATNet_I(Transformer)** | 97.77±0.03 | 30.65±0.60 | 96.26±0.06 | 61.34±0.25 | NA | NA | 96.18±0.07 | 88.52±0.18 |
| | **CATNet_II(Transformer)** | **97.80±0.03** | **33.74±0.23** | **96.26±0.08** | **62.86±0.16** | **NA** | **NA** | **96.31±0.03** | **88.79±0.13** |

Firstly, the proposed CATNet, no matter CATNet-I or CATNet-II, shows stable and outstanding performance and achieves state-of-the-art scores on most metrics with small deviations. The superiority of CATNet over the other methods is greater on the MIMIC-III dataset than on the eICU dataset. In addition, the differences between CATNet in AUPR are much bigger than those in AUC.

**Table 4**
Auc and Aupr of different methods on mortality prediction on the MIMIC-III dataset

| mortality | model | Auc | Aupr |
| --- | --- | --- | --- |
| RNN | DoctorAI | 0.6495±0.0058 | 0.3118±0.0082 |
|  | T_LSTM | 0.5693±0.0272 | 0.2837±0.0194 |
|  | RetainVis | 0.6105±0.0185 | 0.3090±0.0222 |
|  | StageNet | 0.6419±0.0166 | 0.3273±0.0157 |
|  | ConCare | 0.6062±0.0091 | 0.2827±0.0014 |
|  | CATNet_I(GRU) | **0.6562±0.0039** | **0.3612±0.9408** |
|  | CATNet_I(LSTM) | **0.6561±0.0144** | **0.3615±0.0168** |
| Transformer | HiTANet | 0.6660±0.0194 | 0.3760±0.0278 |
|  | CATNet_I(Transformer) | **0.6821±0.0081** | **0.3996±0.0141** |

Secondly, as a method based on basic RNN, the performance of DoctorAI is not outstanding, but stable. It even performs much better than serval RNN variants such as T-LSTM, RetainEx, and ConCare on several prediction tasks. For example, DoctorAI outperforms T-LSTM on all the prediction tasks in Table 2 and Table 3. The reason is that some prediction tasks may depend on many factors. All of them should be comprehensively considered like CATNet. From the results shown in Table 2 and Table 3, we can see that CATNet has a strong task-adaptive ability.

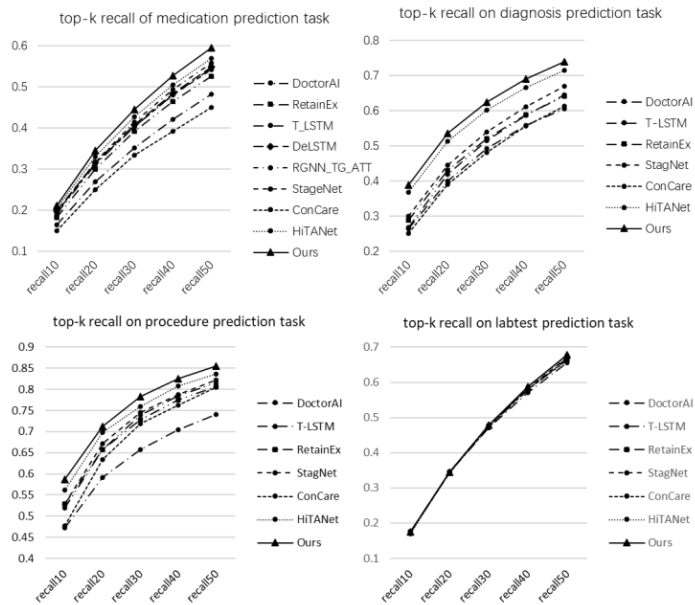

**Fig. 3.** Top-k recalls of different methods for the four tasks listed in Table 2 on the MIMIC-III dataset

Thirdly, comparing the two CATNet methods, CATNet-I shows a little better performance in AUC and much better performance in AUPR than CATNet-II, indicating that the task-aware attention is meaningful for the type of tasks listed in Table 2 and Table 3.

Furthermore, we also show the Top-k (k=10, 20, …, 50) recalls of different methods for the four tasks listed in Table 2 on the MIMIC-III dataset. As shown in Fig. 3, we can see that CATNet outperforms all the other baseline methods consistently on all the tasks, although the difference is small on some tasks (i.e., labtest prediction). This result is consistent with that in Table 2.

### 4.5 Ablation Study

To investigate the importance of each component of CATNet, we compare CATNET_II with its variants that remove some parts of the full CATNet using the same settings as the previous experiments. We still run 10 times to obtain the average performance. The results are shown in Table 5, where "w/o", "Cross", "Vis", "Global" denotes "without", "cross-event attention", "visit-level attention", "global time converter" respectively. It should be noted that we do not remove the "local attention" module from CATNet using Transformer as it is the intrinsic module of Transformer. We can see that all the components contribute to CATNet. From the results, we could conclude that the performance of CATNet decreases when any module is abandoned because the "cross-event attention" module learns correlations among medical events in a visit appropriately, the "local attention" module has the ability to learn relations among historical visits, and the "global time convertor" module is able to model medical event decay rates globally, which simulates the clinical review of the patient historical record to pay attention to each temporal point relatively.

**Table 5**
Auc and Aupr of CATNet and its variants for medication prediction on the MIMIC-III dataset

| model | Auc | Aupr |
| --- | --- | --- |
| CATNet_II(GRU) | **0.8577±0.0010** | **0.4003±0.0031** |
| CATNet_II (GRU) w/o Cross | 0.8470±0.0018 | 0.3742±0.0053 |
| CATNet_II (GRU) w/o Vis | 0.8385±0.0094 | 0.3625±0.0129 |
| CATNet_II(GRU) w/o Global | 0.8424±0.0097 | 0.3663±0.0127 |
| CATNet_II (LSTM) | **0.8584±0.0010** | **0.4023±0.0028** |
| CATNet_II (LSTM) w/o Cross | 0.8491±0.0020 | 0.3772±0.0054 |
| CATNet_II(LSTM) w/o Vis | 0.8426±0.0104 | 0.3687±0.0150 |
| CATNet_II (LSTM) w/o Global | 0.8497±0.0015 | 0.3802±0.0033 |
| CATNet_II(Transformer) | **0.8621±0.0039** | **0.4078±0.0017** |
| CATNet_II (Transformer) w/o Cross | 0.8547±0.0007 | 0.3949±0.0013 |
| CATNet_II(Transformer) w/o Global | 0.8591±0.0012 | 0.4012±0.0040 |

We also conduct an ablation study on each type of medical event, including the "new type" for temporal intervals, and the results are shown in Table 6. In the case of medication prediction, all the other four types of medical events are beneficial to medication prediction. It indicates that the "cross-event attention" module in CATNet has the ability to capture correlations among different events in a visit again, and also has the ability to capture correlations between temporal intervals and medical events.

**Table 6**
Effects of different types of medical events to CATNet for medication prediction on the MIMIC-III dataset.

| Methods | Auc | Aupr |
| --- | --- | --- |
| CATNet_II | **0.8621±0.0039** | **0.4078±0.0017** |
| CATNet_II w/o Diagnosis | 0.8523±0.0008 | 0.3919±0.0019 |
| CATNet_II w/o Labtst | 0.8531±0.0012 | 0.3947±0.0020 |
| CATNet_II w/o Procedure | 0.8544±0.0009 | 0.3952±0.0020 |
| CATNet_II w/o Time | 0.8535±0.0014 | 0.3947±0.0025 |
| CATNet_II w/o Aux | 0.8455±0.0008 | 0.3826±0.0018 |

4.6 *Cross-event Attention Analysis*

From the ablation study and experimental results listed in Table 5, we can conclude that using cross-event attention can significantly improve the performance of medical event prediction. To further illustrate the effectiveness of cross-event attention in CATNet, we conduct case studies to interpret the learned cross-event attention weights and visualize them. Figure 4 shows the attention weights between randomly select three medications and the top 10 diagnoses most related to them. We can see that the cross-event attention can find the strong relationships between the medications and diagnoses. For example, "calcium gluconate" is the conventional medication for "sideroblastic anemia", "ac kidny fail, tubr necr", and "mixed acid-bas bal disorder"

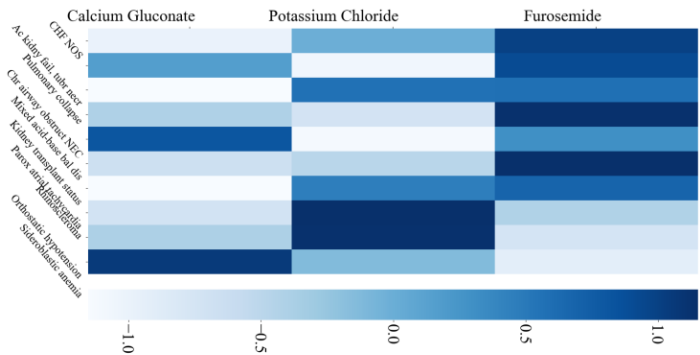

**Fig. 4.** Cross-event attention weights among different events. (diagnoses most related to medications on medication prediction)

## 5. Conclusions

This paper proposes a novel time-aware, event-aware, and task-adaptive deep learning method for medical event prediction, namely, cross-task attention-based time-aware Transformer (CATNet). In CATNet, irregular temporal intervals between neighbor visits are regarded as a new type of medical event and cross-event attention is designed to model correlations among different types of medical events including temporal intervals. The cross-event attention contains two cases corresponding to task-aware and task-unware. Because of the cross-event attention, CATNet is time-aware, event-aware, and task-adaptive. In addition, CATNet also considers each visit at local and global levels. Experiments on two public real-world datasets show the effectiveness of the proposed CATNet and outperform other state-of-the-art methods on various medical event prediction tasks.

## 6. Funding

This paper is supported in part by grants: National Natural Science Foundations of China (U1813215 and 61876052), National Natural Science Foundations of Guangdong, China (2019A1515011158), Guangdong Province Covid-19 Pandemic

## 7. Conflict of interest

The authors declare no conflict of interest.